# Structured Neuron Pruning in Deep Neural Networks Using Multi-Armed Bandits

Salem Ameen and Sunil Vadera

School of Science, Engineering and Environment, University of Salford, Salford, United Kingdom

*Corresponding author: Salem Ameen, s.a.ameen1@salford.ac.uk*

## Abstract

Deep neural networks often contain redundant hidden units. Removing individual weights can reduce the parameter count, but unstructured sparsity is not always easy to exploit in standard dense implementations. This paper develops a structured pruning framework in which complete neurons are removed using multi-armed bandit (MAB) algorithms. Each candidate neuron is treated as an arm, and pulling an arm corresponds to temporarily masking that neuron, measuring the change in loss on a sampled mini-batch, restoring the neuron and updating an estimate of its safe-removal reward. The framework supports stochastic policies, including Epsilon-Greedy, Softmax, UCB1 and Thompson Sampling, and multiplicative-weight policies, including Hedge-style multiplicative weights and EXP3. We evaluate the method on three groups of experiments: tabular classification, tabular regression and deep neural-network benchmarks. The tabular evaluation includes sixteen classification data sets and six regression data sets, while the deep-learning evaluation includes image, text and reasoning benchmarks using LeNet, AlexNet, convolutional networks, LSTM variants, Siamese networks, hierarchical RNNs and end-to-end memory networks. Statistical comparisons using the Friedman test followed by the Nemenyi post-hoc test show significant differences between methods. On the tabular classification tasks, UCB1 obtains the highest mean rank among the pruning policies and improves on the unpruned neural network. On the regression tasks, UCB1 obtains the highest mean rank and is statistically competitive with, or superior to, several standard regression models according to $R^2$. On the deep-learning tasks, UCB1 and Thompson Sampling obtain the strongest ranks, and several MAB policies significantly outperform the unpruned model, magnitude-based neuron pruning and greedy activation-variation pruning. The results show that MAB-based neuron pruning is an effective and computationally practical approach for structured model reduction.

Keywords: neural network pruning; structured pruning; neuron pruning; multi-armed bandits; UCB1; Thompson Sampling; Hedge-style multiplicative weights; EXP3; $R^2$; model compression

## 1. Introduction

Deep neural networks have produced strong results in computer vision, natural language processing, speech recognition and related fields. Their success has often been achieved by increasing the number of layers, neurons and parameters. Large models, however, require more memory, more computation and more energy at inference time. These requirements restrict deployment on embedded devices, mobile systems and real-time applications, where both storage and latency are constrained.

Pruning is a direct way to reduce the size of a trained neural network [1], [2]. The general strategy is to train a model, identify parameters or units that contribute little to prediction, remove them and then retain a smaller model. Weight-level pruning can achieve high sparsity, but the removal of isolated weights often produces irregular sparse matrices [6]. Such sparsity may reduce storage but may not always lead to proportional speedups without specialised sparse computation. Neuron pruning is more structured: removing a neuron removes a whole computational unit together with its associated incoming and outgoing connections. It can therefore reduce dense matrix dimensions and lead to models that are simpler to store and deploy.

The central challenge is to decide which neurons can be removed without damaging predictive performance. A direct method would remove each candidate neuron, test the resulting network on a large sample, rank neurons by the measured loss change and remove the safest ones [3]. This is accurate but expensive. At the other extreme, a simple local statistic such as incoming-weight magnitude or activation variance is cheap but can prune important neurons because it does not directly evaluate the loss after removal [7]-[9]. This paper addresses the trade-off by formulating neuron selection as a fixed-budget MAB problem.

In the proposed formulation, each candidate neuron is an arm. Pulling an arm temporarily removes the corresponding neuron and gives a reward derived from the change in loss. A bandit policy adaptively allocates a limited number of trials: it explores neurons whose importance is uncertain and exploits neurons that repeatedly appear safe to remove. After the play budget is exhausted, the neurons with the highest estimated safe-removal reward are pruned permanently.

This article extends our previous MAB-based weight-pruning framework [1] from unstructured scalar-weight removal to structured neuron pruning. In the earlier work, each arm represented an individual weight, and pulling an arm corresponded to setting that weight to zero. In contrast, the present work treats each candidate neuron as an arm, so that pulling an arm temporarily removes a complete computational unit together with its incoming and outgoing connections. This changes the pruning problem from unstructured sparsification to structured model reduction. The present study also evaluates an extended policy set for neuron pruning, including Softmax/Boltzmann exploration, Hedge-style multiplicative weights and EXP3, and extends the empirical evaluation to tabular classification, tabular regression and deep-learning architectures.

The main contributions are as follows: (i) a formal mathematical formulation of neuron pruning as a fixed-budget MAB ranking problem; (ii) a structured extension of the earlier MAB weight-pruning framework [1] from individual scalar weights to complete neurons with incoming and outgoing connections; (iii) a unified algorithm that supports stochastic and multiplicative-weight bandit policies; (iv) an empirical evaluation on tabular classification, tabular regression and deep-learning benchmarks; (v) statistical comparison using Friedman and Nemenyi tests; and (vi) evidence that UCB1 and Thompson Sampling are particularly effective for structured neuron pruning.

## 2. Related work

Pruning methods differ mainly in how they estimate the importance of a parameter or unit [2], [3]. Magnitude-based methods remove weights, filters or neurons with small norms [6], [8]. They are attractive because they are simple and fast, and they have been successful in large-scale model compression. Their limitation is that a small local magnitude does not necessarily imply that the unit has a small effect on loss.

Sensitivity-analysis methods estimate the effect of perturbing or deleting parameters on the objective function. Optimal Brain Damage uses a second-order Taylor approximation and a diagonal Hessian assumption [4], while Optimal Brain Surgeon retains non-diagonal curvature information [5]. These methods have strong theoretical motivation but are expensive for modern networks because Hessian or inverse-Hessian information can be difficult to compute at scale.

Structured pruning removes complete units such as neurons, filters, channels or feature maps [7]-[9]. This type of pruning is important in practice because it changes the architecture itself rather than only creating sparse weights. Earlier node-pruning methods used relevance or sensitivity scores; later channel and filter pruning methods used activation statistics, filter norms, geometric median measures, optimisation objectives or approximations to loss change. Neuron pruning lies in the same structured-pruning family but focuses on hidden units in dense, recurrent or representation layers.

MAB algorithms provide a principled way of making sequential decisions under uncertainty. Epsilon-Greedy and Softmax policies use simple randomised exploration [11]. UCB methods use optimism in the face of uncertainty and select arms with high empirical reward or high uncertainty [10]. Thompson Sampling uses Bayesian sampling from a posterior reward distribution [12], [13]. Hedge-style multiplicative weights and EXP3 were developed for multiplicative-weight and adversarial-bandit settings [14], [15]. These policies are suitable for pruning because the algorithm must decide which candidate units deserve more evaluation under a limited computation budget.

The proposed method is loss-aware, unlike purely magnitude-based pruning, and avoids Hessian computation, unlike second-order sensitivity methods [4]-[6]. It also differs from brute-force neuron deletion because it uses adaptive sampling rather than exhaustive evaluation [1].

## 3. Mathematical model and pruning algorithms

This section gives the complete mathematical formulation and the full set of neuron-pruning algorithms used in this paper. The formulation follows a fixed-budget bandit setting: each candidate neuron is treated as an arm, pulling an arm means temporarily removing the corresponding neuron, and the observed reward measures whether the temporary removal preserves or improves the loss. At the end of the play budget, the neurons with the largest estimated safe-removal rewards are permanently pruned.

### 3.1 Network and neuron notation

Let

$$D=\{(x_j, y_j): j=1,\dots,N\}$$

be the data set, and let $f_\theta$ be a trained neural network with parameters $\theta$. For a candidate hidden layer $l$, let $K_l$ denote the number of neurons in that layer. Using column-vector activations, the activation vector of layer $l$ is defined as

$$h_l(x)=\sigma_l(W_l h_{l-1}(x)+b_l),$$

where $W_l$ is the weight matrix of layer $l$, $b_l$ is the bias vector, $\sigma_l$ is the activation function, and $h_l(x)$ is a vector with $K_l$ components. More precisely,

$$W_l \in \mathbb{R}^{K_l \times K_{l-1}}, b_l \in \mathbb{R}^{K_l}, h_l(x) \in \mathbb{R}^{K_l}.$$

The candidate arms are the neurons in the selected layer:

$$A_l=\{1,2,\dots,K_l\}.$$

Pruning neuron $i \in A_l$ removes the whole computational unit, not only a single scalar parameter. Under this convention, permanent pruning deletes the incoming parameters of neuron $i$, namely row $i$ of $W_l$ and bias $b_l[i]$. It also deletes the outgoing parameters connected from neuron $i$ to the next layer, namely column $i$ of $W_{l+1}$. In compact notation, pruning neuron $i$ removes

$$W_l[i,:], b_l[i], W_{l+1}[:,i].$$

### 3.2 Structured neuron pruning as masking

During the search phase, temporary neuron removal is represented by a binary mask

$$m \in \{0,1\}^{K_l}.$$

The masked activation is defined by elementwise multiplication:

$$\tilde{h}_l(x;m)=m \odot h_l(x),$$

where $\odot$ denotes elementwise multiplication. Componentwise, this is written as

$$\tilde{h}_{l,i}(x;m)=m_i h_{l,i}(x).$$

This means that the activation of neuron $i$ is set to zero whenever $m_i=0$. The original unpruned network is obtained when all entries of the mask are equal to one:

$$m=1_{K_l},$$

where $1_{K_l}$ is the vector of ones of length $K_l$.

To test the temporary removal of neuron $i$, we use the mask

$$m^{-i} = 1_{K_l} - e_i,$$

where $e_i$ is the $i$-th standard basis vector. The temporarily masked network is denoted by

$$f_\theta(x; m^{-i}).$$

This masking operation is used only to estimate the importance of neurons during the bandit search. After the final pruning set is selected, the corresponding neurons and their associated incoming and outgoing parameters are physically removed from the model.

## 3.3 Loss change after temporary neuron removal

At bandit round $t$, the algorithm samples a mini-batch $B_t$ from the data set $D$. The mini-batch loss under a mask $m$ is defined as:

$$L_{B_t}(m) = \frac{1}{|B_t|} \sum_{(x_j, y_j) \in B_t} \ell(f_\theta(x_j; m), y_j).$$

Here, $\ell(\cdot, \cdot)$ is the task loss. For classification tasks, $\ell$ may be the cross-entropy loss; for regression tasks, it may be the squared-error loss. The unpruned network corresponds to the all-one mask $1_{K_l}$, while the temporary removal of neuron $i$ corresponds to the mask $m^{-i}$.

The loss change caused by temporarily removing neuron $i$ at round $t$ is defined as:

$$\Delta_i(t) = L_{B_t}(1_{K_l}) - L_{B_t}(m^{-i}).$$

This definition follows the same loss-difference principle used in MAB-based weight pruning [1], but applies it to complete neurons rather than individual scalar weights. With this sign convention, $\Delta_i(t) > 0$ means that temporarily removing neuron $i$ reduces the mini-batch loss and therefore improves performance on that mini-batch. If $\Delta_i(t) \approx 0$, the neuron has little measurable effect on the loss. If $\Delta_i(t) < 0$, removing the neuron increases the loss and is likely to harm performance.

The value $\Delta_i(t)$ is therefore used as the basic evidence for estimating how safely neuron $i$ can be removed. A neuron whose temporary removal repeatedly gives a positive or near-zero loss change is treated as a stronger candidate for pruning than a neuron whose removal repeatedly causes a large increase in loss.

## 3.4 Reward functions

The loss change $\Delta_i(t)$ must be converted into a reward before it can be used by a bandit policy [1]. For Epsilon-Greedy, Softmax, UCB1, Hedge-style multiplicative weights and EXP3, we use a bounded reward:

$$r_i(t) = \min\left\{1, \max\left\{0, \frac{\tau + \Delta_i(t)}{c}\right\}\right\}.$$

Here, $\tau \geq 0$ is a tolerance parameter and $c > 0$ is a scaling constant. The tolerance parameter allows a small increase in loss to be accepted when the increase is within a predefined safe range. The scaling constant keeps the reward within the interval $[0, 1]$, which is required by several bandit policies and is especially important for UCB-style confidence bounds.

This reward has the following interpretation. If removing neuron $i$ improves the loss, then $\Delta_i(t)$ is positive and the reward increases. If removal causes only a small loss increase, the reward may still be positive because of the tolerance term $\tau$. If removal causes a large loss increase, the reward is clipped to zero. The outer minimum ensures that the reward never exceeds one.

For Thompson Sampling, we use a binary reward because the method is applied with a Bernoulli success/failure model:

$$b_i(t) = I[\Delta_i(t) \geq -\tau],$$

where $I[\cdot]$ is the indicator function. Thus, $b_i(t)=1$ if the temporary removal of neuron $i$ improves the loss or causes only an acceptable loss increase, and $b_i(t)=0$ otherwise.

Let $n_i(t)$ be the number of times neuron $i$ has been selected up to and including round $t$, and let $\mu_i(t)$ be the empirical mean reward of neuron $i$. After selecting neuron $a_t = i$ and observing reward $r_i(t)$, the empirical mean is updated by:

$$\mu_i(t) = \mu_i(t-1) + \frac{r_i(t) - \mu_i(t-1)}{n_i(t)}.$$

Equivalently,

$$\mu_i(t) = \frac{(n_i(t)-1)\mu_i(t-1) + r_i(t)}{n_i(t)}.$$

For neurons not selected at round $t$, the empirical mean is unchanged:

$$\mu_j(t) = \mu_j(t-1),\; j \neq i.$$

The empirical reward $\mu_i(t)$ estimates the expected safe-removal reward of neuron $i$. After the play budget is exhausted, neurons are ranked according to their empirical rewards, and the neurons with the largest values are selected for permanent structural pruning.

## 3.5 Generic MAB neuron-pruning algorithm

The proposed pruning procedure is formulated as a fixed-budget multi-armed bandit problem. Each candidate neuron in the selected layer is treated as an arm. At each round, the algorithm selects one neuron, temporarily masks it, evaluates the resulting change in mini-batch loss, restores the neuron, and updates the estimated reward associated with that neuron. No neuron is permanently removed during the search phase. Permanent pruning is performed only after the play budget has been exhausted.

Algorithm 1 defines the common pruning framework. Algorithms 2–7 specify the policy-specific selection and update rules for Epsilon-Greedy, Softmax/Boltzmann exploration, UCB1, Thompson Sampling, Hedge-style multiplicative weights, and EXP3. This separation keeps the loss-evaluation step identical across all policies while making the differences between the bandit policies explicit.

**Algorithm 1. Fixed-budget MAB framework for structured neuron pruning**

**Input:** trained network $f_\theta$; data set $D$; candidate layer $l$; candidate neurons $A_l = \{1, \ldots, K_l\}$; play budget $T$; pruning cardinality $q$; mini-batch sampler; reward parameters $\tau$ and $c$; bandit policy $\Pi$.

**Output:** structurally pruned network $f_{\theta'}$.

1. Initialise $n_i \leftarrow 0$ and $\mu_i \leftarrow 0$ for every neuron $i \in A_l$.
2. Initialise the policy-specific state for every neuron $i \in A_l$.

3. For $t=1,2,\ldots,T$, repeat Steps 4–14.
4. Draw a mini-batch $B_t$ from $D$.
5. Select a candidate neuron using the current policy:

$$a_t \leftarrow SELECT\left(\Pi,\{\mu_i,n_i,\zeta_i\}_{i\in A_l},t\right).$$

6. Construct the temporary single-neuron mask:

$$m^{-a_t}=1_{K_l}-e_{a_t}.$$

7. Compute the unmasked mini-batch loss:

$$L_0 \leftarrow L_{B_t}\left(1_{K_l}\right).$$

8. Compute the temporarily masked mini-batch loss:

$$L_1 \leftarrow L_{B_t}\left(m^{-a_t}\right).$$

9. Compute the loss change:

$$\Delta_t \leftarrow L_0-L_1.$$

10. Convert $\Delta_t$ into a reward using the reward rule defined in Section 3.4. For Thompson Sampling, use the corresponding binary reward.
11. Remove the temporary mask; no pruning is permanent during the search phase.
12. Update the number of times the selected neuron has been evaluated:

$$n_{a_t} \leftarrow n_{a_t}+1.$$

13. For policies using bounded rewards, update the empirical mean reward:

$$\mu_{a_t} \leftarrow \mu_{a_t}+\frac{r_t-\mu_{a_t}}{n_{a_t}}.$$

14. Update the policy-specific state using the selected neuron and the observed reward.
15. Select the final pruning set as the $q$ neurons with the largest estimated safe-removal rewards:

$$S_q=Top_q\left(\{\mu_i : i\in A_l\}\right).$$

16. Permanently delete the neurons in $S_q$ and their associated incoming and outgoing parameters.
17. Let $\theta'$ denote the remaining parameter set after deletion.
18. Return the structurally pruned network $f_{\theta'}$.

## 3.6 Epsilon-Greedy neuron pruning

Epsilon-Greedy provides a simple balance between exploration and exploitation [11]. With probability $\epsilon_t$, the policy explores by selecting a neuron uniformly at random. With probability $1-\epsilon_t$, it exploits by selecting the neuron with the largest empirical safe-removal reward. The selection rule is

$$a_t=\begin{cases}\text{a uniformly sampled neuron from } A_l, & \text{with probability } \epsilon_t,\\ \arg\max_{i\in A_l} \mu_i, & \text{with probability } 1-\epsilon_t.\end{cases}$$

A decaying exploration schedule can be used to encourage more exploration at the beginning of the search and more exploitation near the end:

$$\epsilon_t = \epsilon_0 \left( \frac{\epsilon_T}{\epsilon_0} \right)^{t/T},$$

where $\epsilon_0$ is the initial exploration rate and $\epsilon_T$ is the final exploration rate.

**Algorithm 2. Epsilon-Greedy selection policy**

**Input:** empirical rewards $\mu_i$; counts $n_i$; candidate neurons $A_l$; round $t$; play budget $T$; initial and final exploration rates $\epsilon_0$ and $\epsilon_T$.

**Output:** selected neuron $a_t$.

1. If there exists an unplayed neuron $i \in A_l$ with $n_i = 0$, return an unplayed neuron.
2. Set

$$\epsilon_t \leftarrow \epsilon_0 \left( \frac{\epsilon_T}{\epsilon_0} \right)^{t/T}.$$

3. Draw $u_t \sim Uniform(0,1)$.
4. If $u_t < \epsilon_t$, select $a_t$ uniformly at random from $A_l$.
5. Otherwise, select

$$a_t \leftarrow \arg\max_{i \in A_l} \mu_i.$$

6. Return $a_t$.

## 3.7 Softmax/Boltzmann neuron pruning

Softmax, also known as Boltzmann exploration, converts empirical rewards into a probability distribution over the candidate neurons [11]. A neuron with a larger empirical reward is assigned a larger probability of being selected, while lower-scoring neurons can still be explored. The selection probability is

$$P_t(i) = \frac{\exp(\mu_i / \nu_t)}{\sum_{j \in A_l} \exp(\mu_j / \nu_t)},$$

where $\nu_t > 0$ is the temperature. A larger value of $\nu_t$ gives a flatter distribution and encourages exploration. A smaller value concentrates probability on neurons with high empirical rewards. A decaying temperature schedule is

$$\nu_t = \nu_0 \left( \frac{\nu_T}{\nu_0} \right)^{t/T},$$

where $\nu_0$ and $\nu_T$ are the initial and final temperatures.

**Algorithm 3. Softmax/Boltzmann selection policy**

**Input:** empirical rewards $\mu_i$; counts $n_i$; candidate neurons $A_l$; round $t$; play budget $T$; initial and final temperatures $\nu_0$ and $\nu_T$.

**Output:** selected neuron $a_t$.

1. If there exists an unplayed neuron $i \in A_l$ with $n_i = 0$, return an unplayed neuron.
2. Set

$$v_t \leftarrow v_0 \left( \frac{v_T}{v_0} \right)^{t/T}.$$

3. For each neuron $i \in A_l$, compute

$$P_t(i) \leftarrow \frac{\exp(\mu_i / v_t)}{\sum_{j \in A_l} \exp(\mu_j / v_t)}.$$

4. Draw $a_t$ from the categorical distribution defined by $P_t$.
5. Return $a_t$.

For numerical stability, the exponential terms may be computed after subtracting $max_{j \in A_l} \mu_j$ from all empirical rewards. This transformation does not change the resulting probabilities.

## 3.8 UCB1 neuron pruning

UCB1 selects the neuron with the largest upper-confidence score [10]. The score combines the empirical safe-removal reward with an uncertainty bonus. The empirical reward favours neurons that have previously appeared safe to remove, while the uncertainty term favours neurons that have not yet been tested often. For bounded rewards in $[0,1]$, the UCB1 score is

$$U_i(t) = \mu_i + \sqrt{\frac{2 \log t}{n_i}},$$

and the selected neuron is

$$a_t = \arg\max_{i \in A_l} U_i(t).$$

**Algorithm 4. UCB1 selection policy**

**Input:** empirical rewards $\mu_i$; counts $n_i$; candidate neurons $A_l$; round $t$.

**Output:** selected neuron $a_t$.

1. If there exists an unplayed neuron $i \in A_l$ with $n_i = 0$, return an unplayed neuron.
2. For each neuron $i \in A_l$, compute

$$U_i(t) \leftarrow \mu_i + \sqrt{\frac{2 \log t}{n_i}}.$$

3. Select

$$a_t \leftarrow \arg\max_{i \in A_l} U_i(t).$$

4. Return $a_t$.

## 3.9 Thompson Sampling neuron pruning

Thompson Sampling uses a Bayesian success-failure model [12], [13]. For each neuron $i$, let $s_i$ be the number of successful temporary removals and let $f_i$ be the number of failures. A success occurs when temporary removal of the neuron improves the loss or causes only an acceptable loss increase. With a uniform beta prior, the posterior distribution for the success probability of neuron $i$ is

$$p_i \mid s_i, f_i \sim Beta(s_i+1, f_i+1).$$

At each round, Thompson Sampling draws one sample from each posterior and selects the neuron with the largest sampled value:

$$a_t = \arg\max_{i \in A_l} \tilde{p}_i(t), \tilde{p}_i(t) \sim Beta(s_i+1, f_i+1).$$

**Algorithm 5. Thompson Sampling selection and update policy**

**Input:** success counts $s_i$; failure counts $f_i$; candidate neurons $A_l$.

**Output:** selected neuron $a_t$ and updated posterior parameters after reward observation.

1. For each neuron $i \in A_l$, draw
$$\tilde{p}_i(t) \sim Beta(s_i+1, f_i+1).$$
2. Select
$$a_t \leftarrow \arg\max_{i \in A_l} \tilde{p}_i(t).$$
3. Return $a_t$ for temporary masking and reward evaluation.
4. After observing the binary reward $b_t$, update the posterior parameters by setting $s_{a_t} \leftarrow s_{a_t} + b_t$ and $f_{a_t} \leftarrow f_{a_t} + 1 - b_t$.
5. Estimate the safe-removal probability using the posterior mean $\mu_{a_t} \leftarrow (s_{a_t}+1)/(s_{a_t}+f_{a_t}+2)$.

## 3.10 Hedge-style multiplicative-weights neuron pruning

The Hedge-style multiplicative-weights policy maintains a positive selection weight $w_i(t)$ for each neuron $i$ [14]. These quantities are policy weights, not neural-network parameters. The probability of selecting neuron $i$ is

$$P_t(i) = \frac{w_i(t)}{\sum_{j \in A_l} w_j(t)}.$$

After selecting neuron $a_t$ and observing reward $r_t$, the selected policy weight is updated as

$$w_{a_t}(t+1) = w_{a_t}(t) \exp(\eta r_t),$$

where $\eta > 0$ controls how strongly the policy reacts to observed rewards. The unselected policy weights remain unchanged:

$$w_i(t+1) = w_i(t), i \neq a_t.$$

This policy is described as Hedge-style multiplicative weights because the pruning setting observes only the reward of the selected neuron at each round. EXP3, described next, is included as the standard adversarial-bandit multiplicative-weights policy under partial feedback.

**Algorithm 6. Hedge-style multiplicative-weights selection policy**

**Input:** policy weights $w_i > 0$; candidate neurons $A_l$; learning rate $\eta$.

**Output:** selected neuron $a_t$ and updated policy weights after reward observation.

1. For each neuron $i \in A_l$, compute

$$P_t(i) \leftarrow \frac{w_i}{\sum_{j \in A_l} w_j}.$$

2. Draw $a_t$ from the categorical distribution defined by $P_t$.
3. Return $a_t$ for temporary masking and reward evaluation.
4. After observing reward $r_t$, update the selected policy weight:

$$w_{a_t} \leftarrow w_{a_t} \exp(\eta r_t).$$

5. Keep $w_i$ unchanged for all $i \neq a_t$.
6. Normalise the policy weights if required for numerical stability.

## 3.11 EXP3 neuron pruning

EXP3 is an adversarial-bandit policy based on multiplicative weights [15]. It combines the learned probability distribution with uniform exploration. The selection probability is

$$P_t(i) = (1-\gamma)\frac{w_i(t)}{\sum_{j \in A_l} w_j(t)} + \frac{\gamma}{K_l},$$

where $0 < \gamma \leq 1$ controls the amount of forced exploration. Since only the reward of the selected neuron is observed, EXP3 uses an importance-weighted reward estimate:

$$\hat{r}_i(t) = \begin{cases} \frac{r_t}{P_t(a_t)}, & i = a_t, \\ 0, & i \neq a_t. \end{cases}$$

The policy weights are updated using

$$w_i(t+1) = w_i(t) \exp\left(\frac{\gamma \hat{r}_i(t)}{K_l}\right).$$

**Algorithm 7. EXP3 selection and update policy**

**Input:** policy weights $w_i > 0$; candidate neurons $A_l$; exploration parameter $\gamma$.

**Output:** selected neuron $a_t$ and updated policy weights after reward observation.

1. For each neuron $i \in A_l$, compute

$$P_t(i) \leftarrow (1-\gamma)\frac{w_i}{\sum_{j \in A_l} w_j} + \frac{\gamma}{K_l}.$$

2. Draw $a_t$ from the categorical distribution defined by $P_t$.

3. Return $a_t$ for temporary masking and reward evaluation.
4. After observing reward $r_t$, set

$$\hat{r}_{a_t}(t) \leftarrow \frac{r_t}{P_t(a_t)}$$

and

$$\hat{r}_i(t) \leftarrow 0, i \neq a_t.$$

5. For each neuron $i \in A_l$, update

$$w_i \leftarrow w_i \exp\left(\frac{\gamma \hat{r}_i(t)}{K_l}\right).$$

## 3.12 Evaluated bandit policies

The experimental evaluation considers six bandit policies within the same structured neuron-pruning framework: Epsilon-Greedy, Softmax/Boltzmann exploration, UCB1, Thompson Sampling, Hedge-style multiplicative weights and EXP3. These policies were selected to compare the main classes of bandit exploration used in this study: randomised exploration, probability-based exploration, optimistic exploration, Bayesian sampling and multiplicative-weight updating.

This design keeps the pruning operation fixed across all experiments: a candidate neuron is temporarily masked, the mini-batch loss is evaluated, the neuron is restored and the policy state is updated from the observed reward. Therefore, differences in performance are attributable to the selection policy rather than to differences in the pruning mechanism.

Together, these policies provide a representative comparison of MAB-based strategies for fixed-budget structured neuron pruning.

## 3.13 Final structural pruning rule

After the play budget has been exhausted, each candidate neuron has an empirical safe-removal reward. The final pruning set is formed by selecting the $q$ neurons with the largest empirical rewards:

$$S_q = Top_q\left(\{\mu_i\}_{i \in A_l}\right),$$

where $Top_q$ returns the indices of the $q$ largest values among the empirical rewards. The set $S_q$ therefore contains the neurons whose temporary removal most consistently preserved or improved the mini-batch loss during the bandit search.

If $\theta$ denotes the original parameter set, then the pruned parameter set $\theta'$ is obtained by deleting all incoming and outgoing parameters associated with the neurons in $S_q$. Under the layer convention used in this paper, pruning neuron $i$ deletes the incoming row and bias of that neuron and the outgoing column connected to the next layer:

$$W_{l,i,:}, b_{l,i}, W_{l+1,:,i}.$$

Symbolically, the resulting pruned model is denoted by

$$f_{\theta'}.$$

This final step converts temporary masking into a physically smaller deployable model. Temporary masks are used only for estimating neuron importance; the final model is obtained by permanently removing the selected neurons and their associated parameters.

## 3.14 Computational complexity

Let $C_F$ denote the cost of one forward pass on a mini-batch, $T$ the bandit play budget, $K_l$ the number of candidate neurons in layer $l$, $B$ the mini-batch size, and $C_\pi(K_l)$ the cost of selecting an arm under policy $\pi$. Each bandit play requires one forward pass for the unmasked model and one forward pass for the temporarily masked model. Therefore, the total search cost is

$$C_{MAB} = O\left(T\left(2C_F + C_\pi(K_l)\right)\right).$$

For the policies used in this paper, $C_\pi(K_l)$ is linear in the number of candidate neurons under a straightforward implementation:

$$C_\pi(K_l) = O(K_l).$$

Thus, the total search cost can be written as

$$C_{MAB} = O\left(T\left(2C_F + K_l\right)\right).$$

The memory overhead is also linear in the number of candidate neurons because the algorithm stores counts, empirical rewards and policy-specific variables:

$$M_{MAB} = O(K_l).$$

This is substantially cheaper than exhaustive neuron evaluation over the full data set. Since $B$ denotes the mini-batch size, a full-data forward pass costs approximately $(N/B)C_F$. Therefore, evaluating all candidate neurons exhaustively over the full data set costs approximately

$$C_{exh} = O\left(K_l \frac{N}{B} C_F\right).$$

The proposed bandit method is therefore more efficient when the full search cost is much smaller than the exhaustive cost:

$$T\left(2C_F + C_\pi(K_l)\right) \ll K_l \frac{N}{B} C_F.$$

When the forward-pass cost dominates the policy-selection cost, this condition can be approximated as

$$T \ll K_l \frac{N}{B}.$$

In addition, the proposed method avoids the Hessian computation required by second-order pruning methods.

# 4. Experimental methodology

The evaluation consists of three groups of experiments. The first group evaluates feed-forward neural networks on sixteen tabular classification data sets obtained from the UCI Machine Learning Repository and Kaggle. The second group evaluates feed-forward neural networks on six tabular regression data sets using $R^2$ as the main performance metric. The third group evaluates deep neural-network architectures trained on image, text and reasoning tasks. All pruning methods are applied to trained models, and performance is measured on the resulting pruned networks. The results are reported as the empirical evaluation of the proposed MAB-based neuron-pruning method.

### 4.1 Tabular classification data sets and classifiers

The tabular evaluation uses the following data sets: Pima Indians Diabetes, Car Evaluation, Spambase, Adult, Hill-Valley, Titanic, Labeled Faces in the Wild, Wine, Heart Disease, Iris, Abalone, Poker Hand, Glass Identification, Wine Quality, Activity Recognition from Single Chest-Mounted Accelerometer and Lung Cancer. For larger data sets, the data are divided into training, validation and test partitions in the ratio 60:20:20. For smaller data sets, 80% of the data are used for training and 20% for testing; ten-fold cross-validation is used on the training portion for model building and validation.

The neural networks are trained using ReLU activation, a learning rate of 0.001 and 100 epochs. The optimiser, batch size, dropout and weight decay are tuned for each data set. The pruning policies are compared with the unpruned neural network and with a range of standard classifiers, including support vector machines (SVM), linear SVM, k-nearest neighbours, decision trees, bagging, random forests, AdaBoost, Naive Bayes, linear discriminant analysis (LDA), quadratic discriminant analysis (QDA), logistic regression, Gaussian process classifiers, LightGBM and XGBoost. Accuracy, F1 score, precision and recall are used as the principal evaluation metrics [16].

### 4.2 Tabular regression data sets and regressors

The tabular regression evaluation assesses whether MAB-based neuron pruning can preserve predictive performance on continuous-output tasks. Performance is measured using the coefficient of determination, $R^2$. The regression experiments use six data sets: Boston Housing, Airfoil Self-Noise, Auto MPG, Computer Hardware, Concrete Compressive Strength and Parkinsons Telemonitoring. The MAB-pruned neural networks are compared with the original unpruned neural network and with standard regression models, including ordinary least squares, LASSO regression, Bayesian Ridge regression, Kernel Ridge regression, decision tree regression, k-nearest-neighbour regression, support vector regression, AdaBoost, bagging and XGBoost.

The same pruning principle is used as in the classification experiments. Candidate neurons are temporarily removed, the change in validation loss is used to update the bandit reward, and the final pruning decision is made after the play budget has been exhausted. Statistical comparison is carried out using Friedman and Nemenyi tests across the six regression data sets.

### 4.3 Deep-learning experiments

The deep-learning evaluation uses a range of architectures and data sets: a multilayer perceptron (MLP) on Reuters, LeNet on MNIST [20], AlexNet on ImageNet [19], convolutional networks on CIFAR-10, CIFAR-100, IMDB and SVHN, a Siamese network on MNIST, LSTM and bidirectional LSTM models on IMDB, an end-to-end memory network on bAbI and a hierarchical recurrent neural network (RNN) on MNIST. Accuracy is used as the principal metric for the deep-learning experiments.

The baselines are the unpruned model, magnitude-based neuron pruning and a greedy activation-variation pruning method adapted for neurons. The play budget is set to at least twice the number of neurons in the layer being pruned, except where the architecture size requires a larger fixed budget. UCB1 and Thompson Sampling are used without additional policy-specific tuning beyond the common play budget and reward definition. Epsilon-Greedy and Softmax/Boltzmann exploration are evaluated with both fixed and annealed exploration settings. EXP3 uses a fixed exploration parameter $\gamma$, and Hedge-style multiplicative weights uses a fixed learning rate $\eta$.

### 4.4 Statistical comparison

The methods are compared using the Friedman test followed by the Nemenyi post-hoc test [16]-[18]. The Friedman test assesses whether the observed ranks of the methods across data sets differ more than would be expected by chance. Let $K$ be the number of methods and $N$ the number of data sets. If $r_{ij}$ denotes the rank of method $j$ on data set $i$, then the average rank of method $j$ is

$$R_j = \frac{1}{N} \sum_{i=1}^{N} r_{ij}.$$

The Friedman statistic is

$$\chi_F^2 = \frac{12N}{K(K+1)} \left[ \sum_{j=1}^{K} R_j^2 - \frac{K(K+1)^2}{4} \right).$$

When the Friedman test rejects the null hypothesis, the Nemenyi test compares average ranks using the critical difference

$$CD = q_{\alpha,K} \sqrt{\frac{K(K+1)}{6N}}.$$

Here, $q_{\alpha,K}$ is the Nemenyi critical value for $K$ methods at significance level $\alpha$, obtained from the Studentized range distribution. Differences in mean rank larger than $CD$ are considered statistically significant. In this paper, the significance level is set to $\alpha = 0.05$.

## 5. Results

This section reports the empirical results of the proposed MAB-based neuron-pruning methods. The results are organised into tabular classification experiments, tabular regression experiments, deep-learning experiments and statistical comparisons based on Friedman and Nemenyi tests. Higher mean ranks indicate better performance in the ranking tables.

### 5.1 Tabular classification results

The Friedman test indicates significant differences between methods for all four tabular classification metrics. The p-values are $4.49 \times 10^{-15}$ for accuracy, $2.07 \times 10^{-8}$ for F1 score, $3.98 \times 10^{-8}$ for precision and $1.25 \times 10^{-6}$ for recall. These values reject the null hypothesis that all methods perform equivalently across the sixteen tabular data sets. Tables 2a and 2b give the mean ranks; higher ranks indicate better performance.

**Table 2a. Mean ranks on the sixteen tabular classification data sets: highest-ranked methods.**

| Method | Accuracy | F1 | Precision | Recall |
|---|---|---|---|---|
| SVM | 20.218 | 18.844 | 20.156 | 18.656 |
| UCB1 pruning | 19.063 | 18.625 | 16.500 | 18.063 |
| Unpruned NN | 18.531 | 17.843 | 15.593 | 17.219 |
| Gaussian process | 16.906 | 17.344 | 16.781 | 17.219 |
| Thompson Sampling pruning | 16.313 | 16.031 | 13.781 | 15.531 |
| Decay Epsilon-Greedy pruning | 15.719 | 14.531 | 15.094 | 14.313 |
| Epsilon-Greedy pruning | 15.625 | 15.063 | 14.188 | 14.625 |
| Softmax pruning | 15.156 | 13.969 | 14.625 | 13.719 |
| Decay Softmax pruning | 14.969 | 14.531 | 16.625 | 12.719 |
| Random forest | 14.500 | 14.031 | 16.594 | 12.750 |
| Hedge-style MW pruning | 14.406 | 13.531 | 13.938 | 13.031 |
| KNN | 14.156 | 14.438 | 13.125 | 14.312 |
| LightGBM | 14.094 | 12.250 | 12.063 | 10.906 |
| XGBoost | 13.813 | 12.469 | 15.188 | 11.938 |
| Bagging KNN | 13.188 | 11.500 | 13.000 | 11.031 |

**Table 2b. Mean ranks on the sixteen tabular classification data sets: remaining methods.**

| Method | Accuracy | F1 | Precision | Recall |
|---|---|---|---|---|
| EXP3 pruning | 12.281 | 12.344 | 12.094 | 11.563 |
| Decision tree (entropy) | 11.156 | 13.563 | 10.906 | 14.469 |
| Bagging decision tree | 10.407 | 9.719 | 11.906 | 8.125 |
| Decision tree (Gini) | 9.875 | 13.656 | 10.781 | 16.063 |
| QDA | 8.844 | 12.250 | 9.281 | 13.031 |
| Logistic regression | 8.781 | 7.594 | 10.781 | 8.406 |
| LDA | 8.313 | 8.656 | 9.813 | 8.313 |
| Linear SVM | 7.594 | 7.438 | 8.594 | 8.875 |
| AdaBoost | 6.907 | 8.031 | 8.094 | 10.406 |
| Naive Bayes | 4.188 | 7.594 | 5.406 | 9.719 |

**Note.** Hedge-style MW = Hedge-style multiplicative weights.

UCB1 is the strongest pruning policy in the tabular evaluation. It ranks second overall for accuracy, behind only SVM, and it ranks above the unpruned neural network on accuracy, F1 score and recall. Thompson Sampling is also competitive, while Epsilon-Greedy, Decay Epsilon-Greedy, Softmax and Decay Softmax provide useful pruning behaviour across the tabular data sets. Hedge-style multiplicative weights and EXP3 are less consistent in this setting, which may be explained by the fact that the tabular pruning task behaves more like a stochastic best-arm identification problem than a fully adversarial sequence of rewards.

The Nemenyi post-hoc comparisons provide more detailed evidence of where the significant differences occur. For accuracy, the Nemenyi comparison uses a critical difference of CD = 8.066. The post-hoc diagrams for accuracy and F1 score show that MAB-based neuron pruning improves the neural-network model sufficiently to outperform bagging decision trees. For precision, UCB1 pruning outperforms AdaBoost. For recall, the pruned neural network is statistically better than Naive Bayes. These comparisons show that MAB-based neuron pruning is statistically competitive with standard classifiers and can improve the rank of the original neural network after pruning.

## 5.2 Tabular regression results

The regression experiments evaluate the effect of neuron pruning on continuous-output prediction tasks. Performance is measured using $R^2$. The evaluation includes Boston Housing, Airfoil Self-Noise, Auto MPG, Computer Hardware, Concrete Compressive Strength and Parkinsons Telemonitoring. The experiments compare the MAB-pruned neural networks with the original unpruned neural network and with standard regression models.

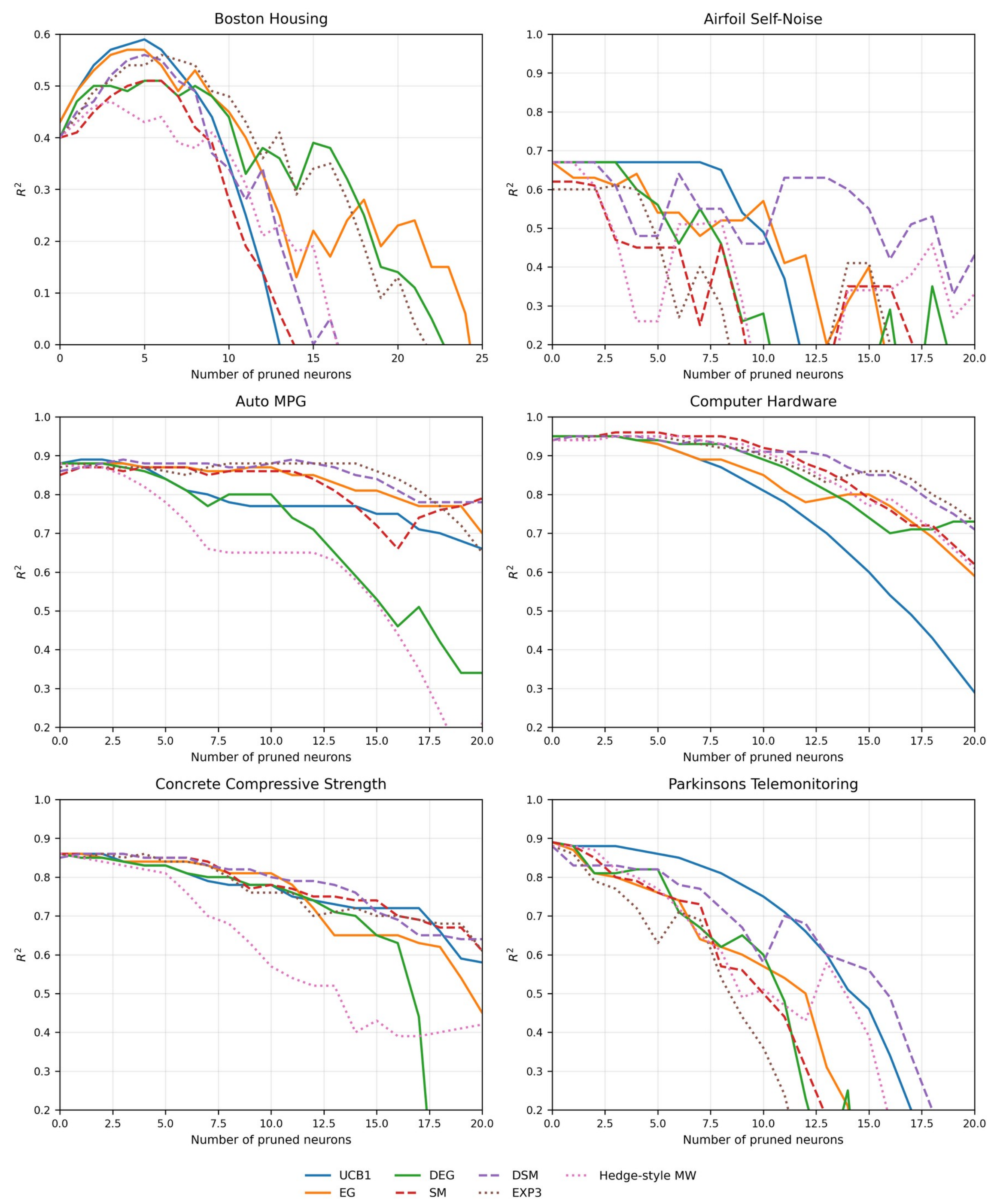

**Figure 5.1. $R^2$ performance of MAB-pruned neural networks as the number of pruned neurons increases across the six regression data sets. The x-axis shows the number of pruned neurons and the y-axis shows $R^2$. In this figure, Hedge denotes the Hedge-style multiplicative-weights policy.**

**Note.** EG = Epsilon-Greedy; DEG = Decay Epsilon-Greedy; SM = Softmax; DSM = Decay Softmax; MW = multiplicative weights. Values are extracted from the saved notebook outputs.

Figure 5.1 shows that the MAB policies have broadly similar behaviour as pruning increases. In general, UCB1 gives the most stable behaviour across the regression data sets, although the relative performance varies with the data set and the number of neurons removed.

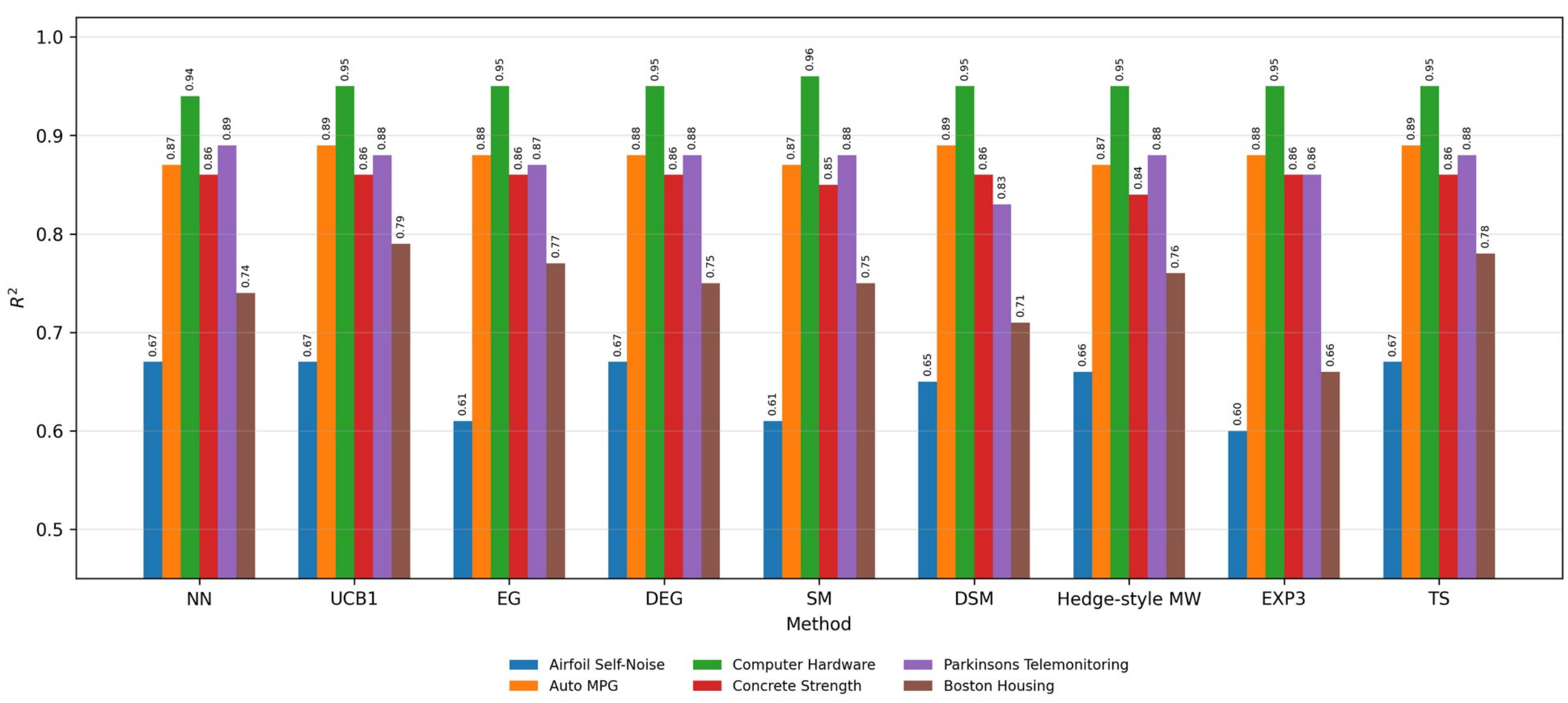

**Figure 5.2. $R^2$ comparison of MAB-pruned neural networks and the original unpruned neural network when approximately 20% of the original network is pruned. In this figure, Hedge denotes the Hedge-style multiplicative-weights policy.**

**Note.** NN = original neural network; EG = Epsilon-Greedy; DEG = Decay Epsilon-Greedy; SM = Softmax; DSM = Decay Softmax; Hedge-style MW = Hedge-style multiplicative weights; TS = Thompson Sampling.

Figure 5.2 compares the MAB-pruned models with the original unpruned neural network. The results show that MAB-based pruning can preserve regression performance and, in several cases, improve on the original model. The strongest behaviour is generally obtained by UCB1 and Thompson Sampling, which remain competitive after pruning.

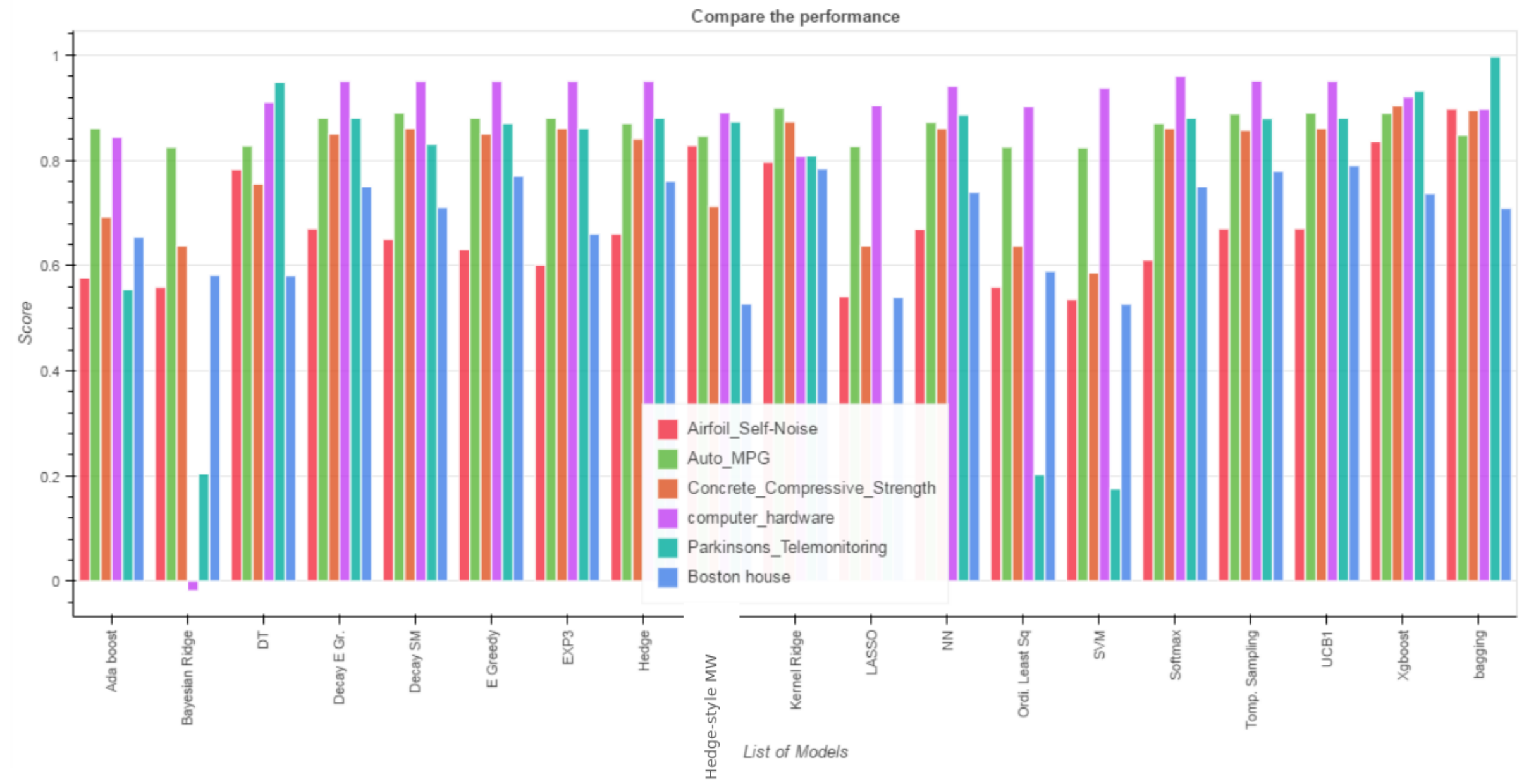

**Figure 5.3. $R^2$ comparison between MAB-pruned neural networks pruned by approximately 25% and standard regression models across the six regression data sets.**

Figure 5.3 compares the pruned models with other regression methods. The results show that the pruned neural networks are competitive with standard regression models, and UCB1 frequently achieves the highest or near-highest $R^2$ performance among the compared methods.

The Friedman test applied to $R^2$ over the six regression data sets and 19 compared algorithms gives $p = 7.45 \times 10^{-7}$, which rejects the null hypothesis that all methods perform equivalently. Table 3 reports the Nemenyi mean ranks; higher ranks indicate better $R^2$ performance.

**Table 3. Mean ranks for the regression experiments based on $R^2$ across six regression data sets.**

| Method | Mean rank |
|---|---|
| UCB1 pruning | 15.500 |
| XGBoost | 15.000 |
| Hedge-style MW pruning | 14.083 |
| Kernel Ridge | 13.000 |
| Bagging | 12.833 |
| Softmax pruning | 12.833 |
| Decay Epsilon-Greedy pruning | 12.750 |
| Unpruned NN | 12.167 |
| Decay Softmax pruning | 12.083 |
| EXP3 pruning | 11.750 |
| Epsilon-Greedy pruning | 11.667 |
| Thompson Sampling pruning | 10.667 |
| Decision tree regression | 9.500 |
| KNN regression | 7.500 |
| AdaBoost regression | 5.500 |
| Ordinary least squares | 4.000 |
| LASSO regression | 3.500 |
| Bayesian Ridge regression | 3.167 |
| SVM regression | 2.500 |

Note. Hedge-style MW = Hedge-style multiplicative weights. Higher rank indicates better $R^2$ performance.

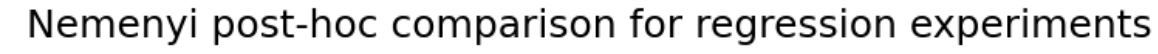
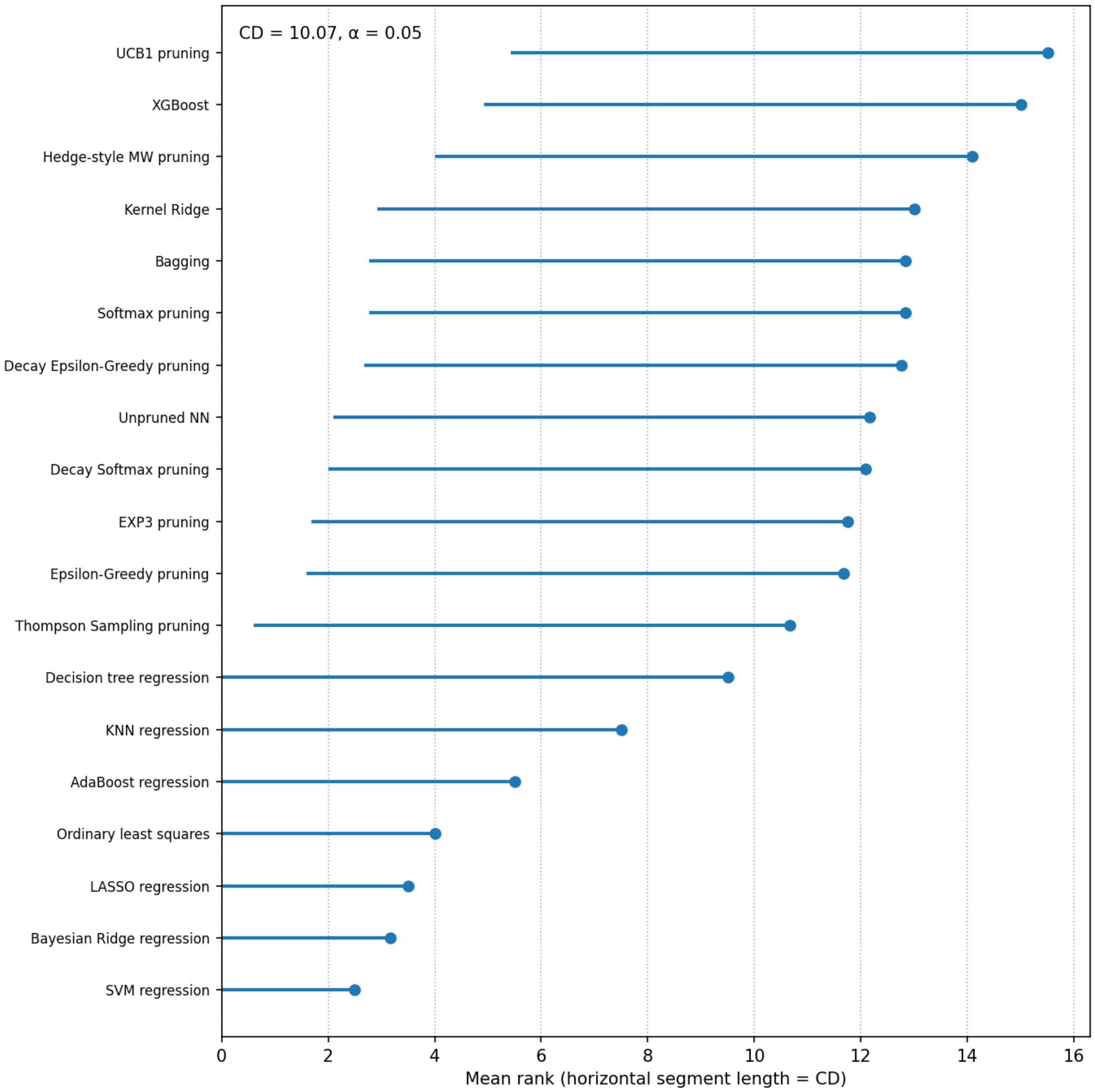

**Figure 5.4. Nemenyi post-hoc comparison for the regression experiments. Dots show mean ranks and horizontal segments have length equal to the Nemenyi critical difference (CD = 10.07, α = 0.05).**

The Nemenyi post-hoc comparison gives a critical difference of CD = 10.07. UCB1 pruning obtains the highest mean rank and is statistically better than ordinary least squares, LASSO regression, Bayesian Ridge regression and SVM regression. Hedge-style multiplicative weights obtains a higher mean rank than the original neural network and is statistically better than ordinary least squares and LASSO regression. Softmax and Decay Epsilon-Greedy also obtain higher mean ranks than the original unpruned model and are statistically better than SVM regression. These results show that the proposed MAB-based neuron-pruning framework is not limited to classification; it is also effective for regression tasks measured using $R^2$.

### 5.3 Deep-learning results

Table 4 reports the original accuracy, the best MAB result, the policy or policies achieving that result, the two non-bandit pruning baselines and the percentage of neurons pruned. The full policy-by-policy accuracy results are provided in Appendix A.

Table 4a. Summary of the deep-learning neuron-pruning results: feed-forward and convolutional models.

| Architecture | Data | Layer | Neurons | Original | Best MAB | Best MAB policy/policies | Magnitude | Greedy | Pruned |
|---|---|---|---|---|---|---|---|---|---|
| MLP | Reuters | 2 | 512 | 0.786 | 0.790 | EG, DEG, SM, DSM, UCB1, TS, Hedge-style MW, EXP3 | 0.750 | 0.740 | 50% |
| LeNet | MNIST | 3 | 128 | 0.980 | 0.990 | EG, DEG, SM, DSM, UCB1, TS, Hedge-style MW, EXP3 | 0.970 | 0.950 | 62% |
| AlexNet FC6 | ImageNet | 6 | 4096 | 0.560 | 0.580 | UCB1, TS | 0.530 | 0.570 | 50% |
| AlexNet FC7 | ImageNet | 7 | 4096 | 0.560 | 0.580 | EG, DEG | 0.550 | 0.540 | 50% |
| ConvNet | CIFAR-10 | 5 | 512 | 0.810 | 0.820 | EG, DEG, SM, DSM, UCB1, TS, Hedge-style MW, EXP3 | 0.810 | 0.800 | 21% |
| ConvNet | CIFAR-100 | 5 | 512 | 0.430 | 0.450 | EG, DEG, UCB1, TS | 0.430 | 0.410 | 20% |
| ConvNet | IMDB | 4 | 250 | 0.870 | 0.890 | EG, DEG, DSM, UCB1, TS, Hedge-style MW | 0.830 | 0.840 | 20% |
| ConvNet | SVHN | 5 | 512 | 0.960 | 0.970 | EG, DEG, SM, DSM, UCB1, TS, Hedge-style MW, EXP3 | 0.900 | 0.920 | 39% |

Table 4b. Summary of the deep-learning neuron-pruning results: Siamese, recurrent and memory models.

| Architecture | Data | Layer | Neurons | Original | Best MAB | Best MAB policy/policies | Magnitude | Greedy | Pruned |
|---|---|---|---|---|---|---|---|---|---|
| Siamese network | MNIST | 2 | 128 | 0.990 | 1.000 | EG, UCB1, TS | 0.930 | 0.930 | 50% |
| LSTM | IMDB | 2 | 128 | 0.800 | 0.810 | EG, DEG, SM, DSM, UCB1, TS, Hedge-style MW, EXP3 | 0.790 | 0.780 | 23% |
| Bidirectional LSTM 1 | IMDB | 2 | 64 | 0.820 | 0.840 | EG, UCB1, TS | 0.800 | 0.810 | 78% |
| Bidirectional LSTM 2 | IMDB | 3 | 64 | 0.820 | 0.830 | EG, DEG, SM, DSM, UCB1, TS, Hedge-style MW | 0.800 | 0.800 | 78% |
| End-to-end memory net | bAbI | 4 | 32 | 0.860 | 0.880 | DEG, UCB1, EXP3 | 0.810 | 0.830 | 40% |
| Hierarchical RNN | MNIST | 2 | 128 | 0.960 | 0.980 | SM, UCB1, Hedge-style MW, EXP3 | 0.900 | 0.920 | 15% |

**Note.** EG = Epsilon-Greedy; DEG = Decay Epsilon-Greedy; SM = Softmax; DSM = Decay Softmax; TS = Thompson Sampling; Hedge-style MW = Hedge-style multiplicative weights. Multiple policies indicate that the same best MAB accuracy was achieved by more than one policy.

The results indicate that MAB-based policies generally preserve or improve accuracy while removing a substantial proportion of neurons. On LeNet/MNIST, pruning 62% of the selected layer improves accuracy from 0.98 to 0.99. On AlexNet/ImageNet, pruning 50% of FC6 or FC7 preserves or improves Top-1 accuracy for the strongest MAB policies. On CIFAR-10, CIFAR-100, IMDB and SVHN convolutional models, MAB pruning improves or maintains performance and outperforms the two non-bandit pruning baselines. On recurrent models, UCB1 and Thompson Sampling again perform strongly, and MAB pruning remains effective even when pruning a high proportion of neurons in bidirectional LSTM layers.

## 5.4 Statistical comparison on deep-learning tasks

The Friedman test over the deep-learning experiments gives $p = 5.597 \times 10^{-17}$, rejecting the null hypothesis that all pruning methods have equivalent performance. The Nemenyi post-hoc comparison gives a critical difference of CD = 3.218. Table 5 gives the Nemenyi mean ranks; higher ranks indicate better performance. Figure 5.5 provides the corresponding critical-difference comparison.

Table 5. Mean ranks for deep-learning neuron-pruning experiments.

| Method | Mean rank |
|---|---|
| UCB1 | 8.679 |
| Thompson Sampling | 8.142 |
| Epsilon-Greedy | 8.107 |
| Decay Epsilon-Greedy | 7.607 |
| Decay Softmax | 6.964 |
| Hedge-style MW | 6.714 |
| Softmax | 6.643 |
| EXP3 | 6.036 |
| Unpruned model | 3.321 |
| Greedy prune | 2.034 |
| Magnitude prune | 1.750 |

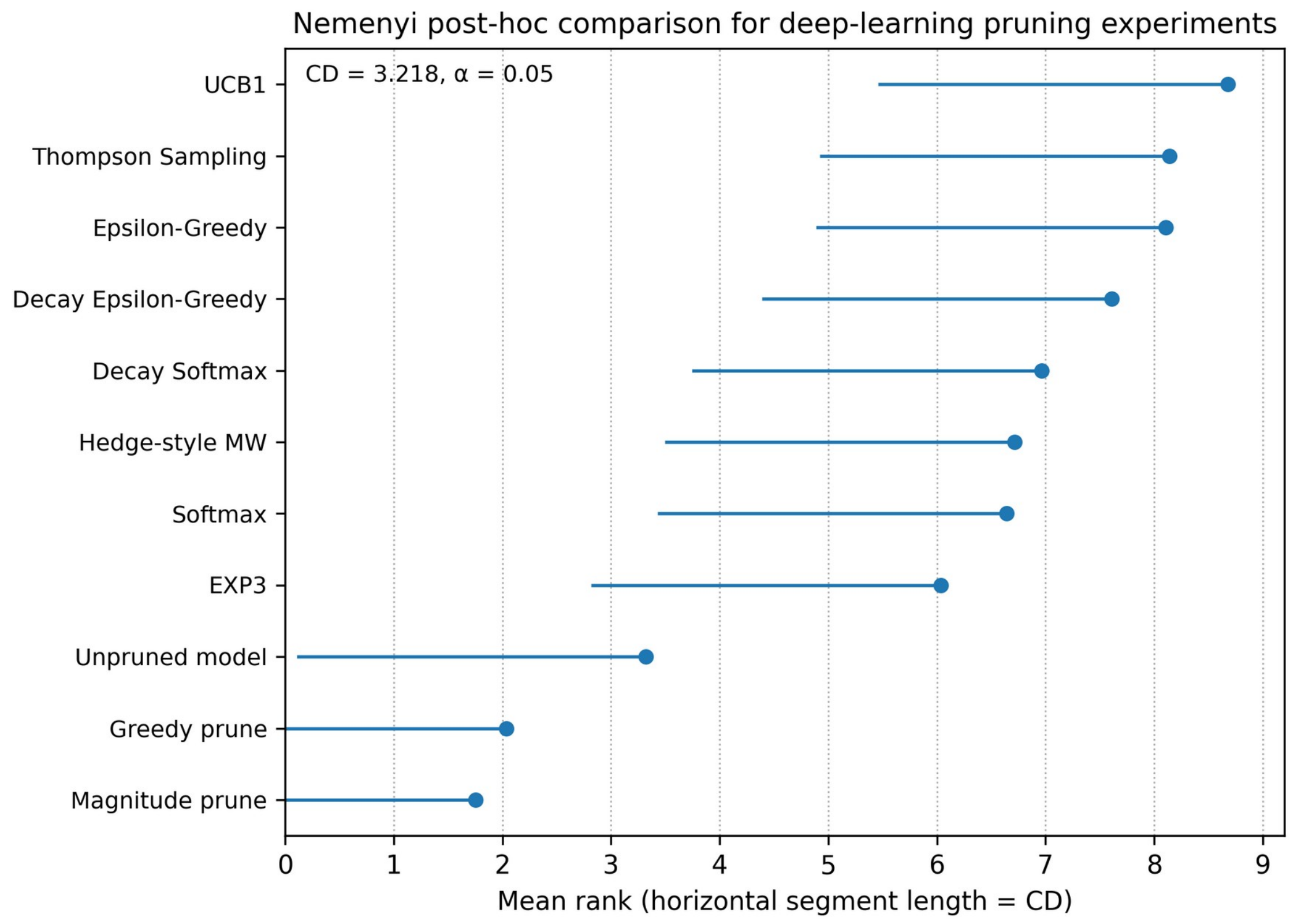

**Figure 5.5. Nemenyi post-hoc comparison for the deep-learning neuron-pruning experiments. Dots show mean ranks, and the horizontal segments have length equal to the Nemenyi critical difference (CD = 3.218, α = 0.05).**

UCB1 obtains the highest mean rank, followed by Thompson Sampling, Epsilon-Greedy and Decay Epsilon-Greedy. UCB1, Thompson Sampling, Epsilon-Greedy, Decay Epsilon-Greedy, Decay Softmax, Hedge-style multiplicative weights and Softmax exceed the original unpruned model by more than the Nemenyi critical difference. EXP3 ranks above the unpruned model but does not exceed it by the critical difference. Decay Softmax, Hedge-style multiplicative weights and Softmax also exceed greedy pruning and magnitude-based pruning by more than the critical difference. The non-bandit pruning baselines, namely magnitude-based neuron pruning and greedy activation-variation pruning, obtain the lowest ranks.

These results support the conclusion that adaptive loss-aware bandit sampling provides a stronger criterion for structured neuron pruning than static magnitude-based or greedy activation-variation baselines. The strongest evidence is for UCB1 and Thompson Sampling, which combine strong empirical performance with a principled mechanism for balancing exploration and exploitation under a fixed pruning budget.

## 6. Discussion

The experimental evidence shows that structured neuron pruning is well suited to a bandit formulation. Estimating neuron importance directly is expensive because the effect of removing a neuron depends on the data sample, the selected layer and the remaining network. Bandit algorithms provide a practical compromise: they do not evaluate every neuron exhaustively over the full data set, but they also do not rely only on local statistics such as weight magnitude or activation variation. Instead, they repeatedly evaluate temporary neuron removal and allocate more trials to candidates that appear promising under the observed loss changes.

UCB1 performs particularly well because it combines empirical safe-removal reward with an uncertainty bonus. At the beginning of the pruning process, the uncertainty term encourages exploration of candidate neurons. As evidence accumulates, the policy increasingly favours neurons whose temporary removal consistently produces little or no loss increase. This behaviour is appropriate for fixed-budget pruning, where the objective is not only to find one removable neuron but to obtain a reliable ranking of neurons for final structural pruning. Thompson Sampling also performs strongly because it maintains uncertainty through a posterior distribution and can continue sampling plausible pruning candidates without committing too early to a fixed ranking.

The results also show that simple stochastic policies can be effective. Epsilon-Greedy and Decay Epsilon-Greedy achieve strong ranks in the deep-learning experiments, indicating that a controlled amount of random exploration can be sufficient when the reward signal is informative. Softmax and Decay Softmax also perform competitively in several cases because they assign higher probability to neurons with larger empirical safe-removal rewards while still allowing lower-ranked neurons to be explored.

Hedge-style multiplicative weights and EXP3 provide useful comparisons with multiplicative-weight selection methods. EXP3 is designed for adversarial bandit feedback, while Hedge-style multiplicative weights provides a related policy-weighting mechanism adapted here to the pruning setting. Their performance is generally less competitive than UCB1 and Thompson Sampling, although they still outperform static non-bandit pruning baselines in several comparisons. This may be because the pruning process used here is noisy but not fully adversarial: during the scoring phase the network is fixed, each selected neuron is restored after temporary removal, and rewards are estimated from sampled mini-batches. Under these conditions, stochastic bandit policies that exploit empirical reward and uncertainty appear better matched to the pruning task.

Compared with weight pruning, neuron pruning is more structured. Removing a single weight affects one scalar connection, whereas removing a neuron removes an entire computational unit together with its incoming and outgoing connections. This can reduce the effective width of a layer and produce a smaller dense architecture that is easier to deploy than an irregular sparse model. At the same time, neuron pruning is a higher-risk operation because an important neuron may influence many downstream activations. The loss-aware reward is therefore essential: it evaluates the effect of temporary removal on model performance before any permanent deletion is made.

The results also show that pruning can sometimes improve predictive performance. This is consistent with the view that over-parameterised neural networks may contain redundant neurons that contribute little to prediction or may increase overfitting. Removing such units can act as a form of structural regularisation. The improvement is not universal, and the safe pruning level depends on the architecture, layer and data set. However, the results show that MAB-based neuron pruning can remove a substantial proportion of neurons while preserving, and in several cases improving, predictive performance.

The regression experiments further support this conclusion. Using $R^2$ as the evaluation metric, UCB1 obtains the highest mean rank across the six regression data sets and is statistically better than several standard regression models. This shows that the proposed reward formulation is not restricted to classification accuracy; it can also be applied to continuous-output prediction tasks by measuring the effect of temporary neuron removal on the regression loss.

Overall, the discussion supports the central finding of the paper: adaptive loss-aware bandit sampling provides a stronger criterion for structured neuron pruning than static magnitude-based or greedy activation-variation baselines. Among the evaluated policies, UCB1 and Thompson Sampling provide the strongest overall balance between exploration, exploitation and pruning reliability.

## 7. Threats to validity

The evaluation covers a range of tabular data sets, deep-learning architectures and pruning policies, but the results are still conditioned on the selected architectures, layers, hyperparameters and play budgets. Different layer choices, pruning percentages or play budgets may change the relative performance of the policies. The reward is estimated from sampled mini-batches, which introduces sampling variability; the repeated bandit plays reduce this effect, but the estimated reward remains an approximation of the true effect of removing a neuron.

Accuracy is the main metric in the deep-learning experiments, while $R^2$ is used for the regression experiments. Although the percentage of pruned neurons is reported, a more complete deployment analysis would also include parameter reduction, FLOPs, inference latency, memory usage and energy consumption on target hardware. The experiments focus on pruning after training and do not include systematic fine-tuning after pruning. Combining MAB-based neuron pruning with fine-tuning or iterative prune-and-retrain cycles may yield additional compression or accuracy recovery. Finally, the evaluation focuses on selected feed-forward, convolutional, recurrent and memory-network architectures; further work is needed to assess the method on transformer-based models and larger modern architectures.

## 8. Conclusion and future work

This paper presented a structured neuron-pruning method based on multi-armed bandits. Each candidate neuron is treated as an arm, and temporary neuron removal is used to estimate a safe-removal reward from the change in mini-batch loss. The same pruning framework supports stochastic policies such as Epsilon-Greedy, Softmax, UCB1 and Thompson Sampling, as well as multiplicative-weight policies including Hedge-style multiplicative weights and EXP3.

The empirical evaluation shows that MAB-based neuron pruning can remove substantial proportions of neurons from selected layers while preserving, and in several cases improving, predictive performance. On the tabular classification data sets, UCB1 obtains the highest mean rank among the pruning policies and improves the rank of the unpruned neural network. On the regression data sets, UCB1 obtains the highest mean rank for $R^2$ and is statistically better than several standard regression models according to the Nemenyi post-hoc test.

On the deep-learning benchmarks, UCB1 and Thompson Sampling are the strongest overall methods, and the Friedman and Nemenyi tests show statistically significant differences between the proposed policies, the unpruned model and the non-bandit pruning baselines.

The results support the conclusion that adaptive loss-aware bandit sampling is an effective criterion for structured neuron pruning. Future work will investigate iterative pruning with fine-tuning, joint pruning across multiple layers, and hardware-aware reward functions that incorporate parameter count, FLOPs, latency, memory usage or energy consumption. Another promising direction is to combine neuron-level and feature-map-level bandit pruning so that both model size and inference time are reduced within a single adaptive framework. Extensions to transformer architectures are also an important direction for future research.

## Declarations

**Conflict of interest:** The authors declare no conflict of interest.

**Funding:** No specific external funding is declared for this manuscript.

**Data and code availability:** The implementation materials associated with the pruning experiments are available at https://github.com/SalemAmeen?page=6&tab=repositories

**Author contributions**: Salem Ameen developed the algorithms, implemented the experiments and drafted the manuscript. Sunil Vadera supervised the research, contributed to the design of the study and reviewed the manuscript.

## Appendix A. Full deep-learning accuracy results

Tables A1 and A2 provide the full accuracy results for the deep-learning experiments. The columns are divided across two tables to improve presentation.

**Table A1. Original and stochastic MAB accuracy results.**

| Model/data | Original | E-greedy | Decay E | Softmax | Decay SM | UCB1 | TS |
|---|---|---|---|---|---|---|---|
| MLP Reuters | 0.786 | 0.790 | 0.790 | 0.790 | 0.790 | 0.790 | 0.790 |
| LeNet MNIST | 0.980 | 0.990 | 0.990 | 0.990 | 0.990 | 0.990 | 0.990 |
| AlexNet FC6 ImageNet | 0.560 | 0.570 | 0.570 | 0.550 | 0.550 | 0.580 | 0.580 |
| AlexNet FC7 ImageNet | 0.560 | 0.580 | 0.580 | 0.560 | 0.560 | 0.570 | 0.570 |
| ConvNet CIFAR-10 | 0.810 | 0.820 | 0.820 | 0.820 | 0.820 | 0.820 | 0.820 |
| ConvNet CIFAR-100 | 0.430 | 0.450 | 0.450 | 0.440 | 0.450 | 0.450 | 0.450 |
| ConvNet IMDB | 0.870 | 0.890 | 0.890 | 0.880 | 0.890 | 0.890 | 0.890 |
| ConvNet SVHN | 0.960 | 0.970 | 0.970 | 0.970 | 0.970 | 0.970 | 0.970 |
| Siamese MNIST | 0.990 | 1.000 | 0.990 | 0.990 | 0.990 | 1.000 | 1.000 |
| LSTM IMDB | 0.800 | 0.810 | 0.810 | 0.810 | 0.810 | 0.810 | 0.810 |
| BiLSTM 1 IMDB | 0.820 | 0.840 | 0.830 | 0.830 | 0.830 | 0.840 | 0.840 |
| BiLSTM 2 IMDB | 0.820 | 0.830 | 0.830 | 0.830 | 0.830 | 0.830 | 0.830 |
| End-to-end memory bAbI | 0.860 | 0.870 | 0.880 | 0.870 | 0.870 | 0.880 | 0.870 |
| Hierarchical RNN MNIST | 0.960 | 0.970 | 0.950 | 0.980 | 0.970 | 0.980 | 0.970 |

**Table A2. Adversarial policies, non-bandit baselines and pruning percentages.**

| Model/data | Hedge-style MW | EXP3 | Magnitude | Greedy | Pruned |
|---|---|---|---|---|---|
| MLP Reuters | 0.790 | 0.790 | 0.750 | 0.740 | 50% |
| LeNet MNIST | 0.990 | 0.990 | 0.970 | 0.950 | 62% |
| AlexNet FC6 ImageNet | 0.530 | 0.540 | 0.530 | 0.570 | 50% |
| AlexNet FC7 ImageNet | 0.560 | 0.560 | 0.550 | 0.540 | 50% |
| ConvNet CIFAR-10 | 0.820 | 0.820 | 0.810 | 0.800 | 21% |
| ConvNet CIFAR-100 | 0.440 | 0.440 | 0.430 | 0.410 | 20% |
| ConvNet IMDB | 0.890 | 0.880 | 0.830 | 0.840 | 20% |
| ConvNet SVHN | 0.970 | 0.970 | 0.900 | 0.920 | 39% |
| Siamese MNIST | 0.990 | 0.990 | 0.930 | 0.930 | 50% |
| LSTM IMDB | 0.810 | 0.810 | 0.790 | 0.780 | 23% |
| BiLSTM 1 IMDB | 0.830 | 0.800 | 0.800 | 0.810 | 78% |
| BiLSTM 2 IMDB | 0.830 | 0.800 | 0.800 | 0.800 | 78% |
| End-to-end memory bAbI | 0.870 | 0.880 | 0.810 | 0.830 | 40% |
| Hierarchical RNN MNIST | 0.980 | 0.980 | 0.900 | 0.920 | 15% |